\documentclass[conference]{IEEEtran}

\IEEEoverridecommandlockouts

\usepackage{cite}
\usepackage{amsmath,amssymb,amsfonts}
\usepackage{algorithmic}
\usepackage{graphicx}
\usepackage{textcomp}
\usepackage{xcolor}
\usepackage{lettrine}
\usepackage{fix-cm}
\usepackage{booktabs}
\usepackage{multirow}
\usepackage{siunitx}
\usepackage{makecell}
\usepackage{tabularx}
\def\BibTeX{{\rm B\kern-.05em{\sc i\kern-.025em b}\kern-.08em
    T\kern-.1667em\lower.7ex\hbox{E}\kern-.125emX}}
\begin{document}

\title{Cross-Stage Attention Multi-Expert Network for Radiologist-Inspired Breast Ultrasound Diagnosis\\
{\footnotesize \textsuperscript{}}
\thanks{}
}

\author{
\IEEEauthorblockN{Xinyang Zhai}
\and
\IEEEauthorblockN{Chong Yang*}
*Corresponding author
\and
\IEEEauthorblockN{Ruizhi Zhang}
}

\maketitle

\begin{abstract}
Breast ultrasound imaging is an important noninvasive method for early breast cancer diagnosis, but automatic benign/malignant classification remains challenging due to tumor heterogeneity, blurred boundaries, and data imbalance. To improve feature representation and classification accuracy, this paper proposes the Cross-Stage Attention Mixture-of-Experts Network (CSA-MoE-Net). It adopts a Cross-Stage Attention-enhanced ResNet-18 as the backbone, in which the Cross-Stage Attention module adaptively recalibrates multi-level features, thereby enhancing key tumor features and suppressing redundancy. A three-branch Mixture of Experts (MoE) Block learns complementary features from the Whole Tumor Image, Tumor Core, and Boundary, and an Adaptive Gating Network fuses them to capture morphological, textural, and contextual information. The fused features are denoted as Fused Expert Feature (FEF) in the architecture.
Experiments on a balanced dataset of 2,129 breast ultrasound images show that, averaged over 20 independent runs, the model achieves an accuracy of 96.33\%, precision of 94.09\%, recall of 98.53\%, F1-score of 96.25\%, and AUC of 99.50\%. Compared to the baseline ResNet-18, these metrics improve by 3.01, 0.70, 5.37, 2.98, and 5.42 percentage points, respectively. The proposed mechanism requires no invasive modification and can be seamlessly embedded into VGG-16, DenseNet-121, etc., yielding stable performance gains, thus providing reliable support for computer-aided diagnosis.
\end{abstract}


\section{Introduction}
\lettrine[lines=2, lhang=0, findent=0.2em]{B}{reast} Cancer has become the most common malignant tumor among women worldwide. According to the latest 2025 statistics from the International Agency for Research on Cancer (IARC) of the World Health Organization, new cases have surpassed lung cancer, making it the leading global malignant tumor. This poses a serious threat to human life and health and has become a major public health issue requiring urgent attention [1]. Due to the insidious onset and atypical clinical manifestations of early breast cancer, accurate benign-malignant differentiation of tiny lesions, blurred-boundary lesions, and highly heterogeneous tumors in ultrasound imaging remains a key challenge in clinical and medical imaging fields [2].

Deep learning-based intelligent diagnosis technology for breast ultrasound has evolved over more than twenty years, gradually moving from laboratory research to clinical application, with four distinct stages [2]. From 2000 to 2012, the traditional machine learning stage relied on manually designed feature extraction operators to capture texture, morphology, gray level, and boundary features of lesions. Combined with models like support vector machines, random forests, and logistic regression, these methods were highly dependent on radiologists’ expertise, had poor generalization, and struggled to capture complex nonlinear lesion features [3]. From 2012 to 2017, the deep learning initiation stage saw convolutional neural networks (AlexNet, VGG, ResNet) enable end-to-end automatic feature extraction [4]. Transfer learning migrated pre-trained weights from natural images to breast ultrasound datasets, alleviating sample scarcity and achieving classification accuracy significantly better than traditional methods [5]. From 2018 to 2020, the field entered rapid development: segmentation models (U-Net) and detection algorithms (Faster R-CNN, YOLO) were widely applied, enabling integrated modeling of lesion detection, segmentation, and classification [6]. Multi-scale feature fusion and attention mechanisms enhanced the focus on key areas, with some models achieving diagnostic performance comparable to that of experienced radiologists on single-center datasets [7]. Since 2021, focus has shifted to clinical implementation and refinement, expanding beyond classification accuracy to clinical utility (BI-RADS grading, axillary lymph node metastasis assessment, neoadjuvant chemotherapy efficacy prediction, molecular subtyping) [8]. Explainable AI, few-shot learning, weakly supervised learning, and federated learning have addressed medical data privacy, scarcity, and model “black-box” issues [9], with multiple AI-assisted diagnostic products approved by China’s NMPA [10].

Despite significant progress, three core clinical bottlenecks remain. First, architectural design is disconnected from clinical logic: early single-branch, whole-image input networks achieved end-to-end classification but failed to fully mine discriminative lesion information [11]. Later studies improved local features by manually cropping lesions and combining with global images, but manual intervention caused cumbersome processes and inconsistent results [12]. Second, complex lesion recognition is inadequate: mainstream CNNs achieve only 82.3\%–86.7\% average accuracy and 0.85–0.89 AUC on public datasets like BUSI [13]. For poorly defined infiltrating lesions, accuracy is generally below 70\% [14], and recall for lesions $\leq$5mm is only 0.65 [15], highlighting shortcomings of traditional single-branch global CNNs in extracting features from blurry and tiny lesions. Third, technical mechanisms have inherent limitations: while Mixture of Experts (MoE) architectures alleviate single-network adaptability issues via multi-expert collaboration, existing work divides experts by abstract feature types rather than anatomical regions, reducing model interpretability [16].

In CNN optimization, residual connections in ResNet solved deep network degradation but did not address differentiated channel importance allocation [4]. SENet introduced channel attention to enhance feature weighting and adaptive recalibration, but only worked at single layers without cross-layer connections [17]. DenseNet proposed cross-layer connections but forced the concatenation of all shallow features with fixed patterns and high redundancy [18]. Recent Layer Attention methods dynamically weight multi-layer features, reducing training loss by 15\%–30\% and improving downstream accuracy by 2\%–5\% in large language models, which provides valuable insights and good inspiration for this study [19,20].

To address these shortcomings—insufficient single-branch feature expression, unreasonable MoE expert division, and limitations of traditional attention—this paper proposes the Cross-Stage Attention Mixture-of-Experts Network (CSA-MoE-Net), with two targeted improvements.

\begin{itemize}
\item First, inspired by the Attention Residual architecture’s cross-layer information interaction [21], a Cross-Stage Attention-enhanced ResNet-18 backbone is designed, breaking the limitations of single-layer attention, establishing dynamic feature associations across four residual stages, and using learnable query vectors for hierarchical importance evaluation to integrate shallow details and deep semantic features. 

\item Second, a clinically driven three-expert MoE architecture is constructed, aligning model decisions with radiologists’ “Whole Tumor Image – Tumor Core – Boundary analysis” logic via three complementary branches: Expert-Img, Expert-Tumor, and Expert-Boundary. 

\item Third, the proposed CSA-MoE-Net achieves excellent performance in intelligent breast ultrasound diagnosis, with high diagnostic accuracy and strong generalization ability that meet clinical application requirements.
\end{itemize}
In conclusion, this model aims to build a multi-branch, specialized architecture consistent with clinical thinking, enhance generalization and clinical utility, and provide a new technical solution for intelligent breast ultrasound diagnosis.

\begin{figure*}[!t]
\centering
\includegraphics [width=1\linewidth]{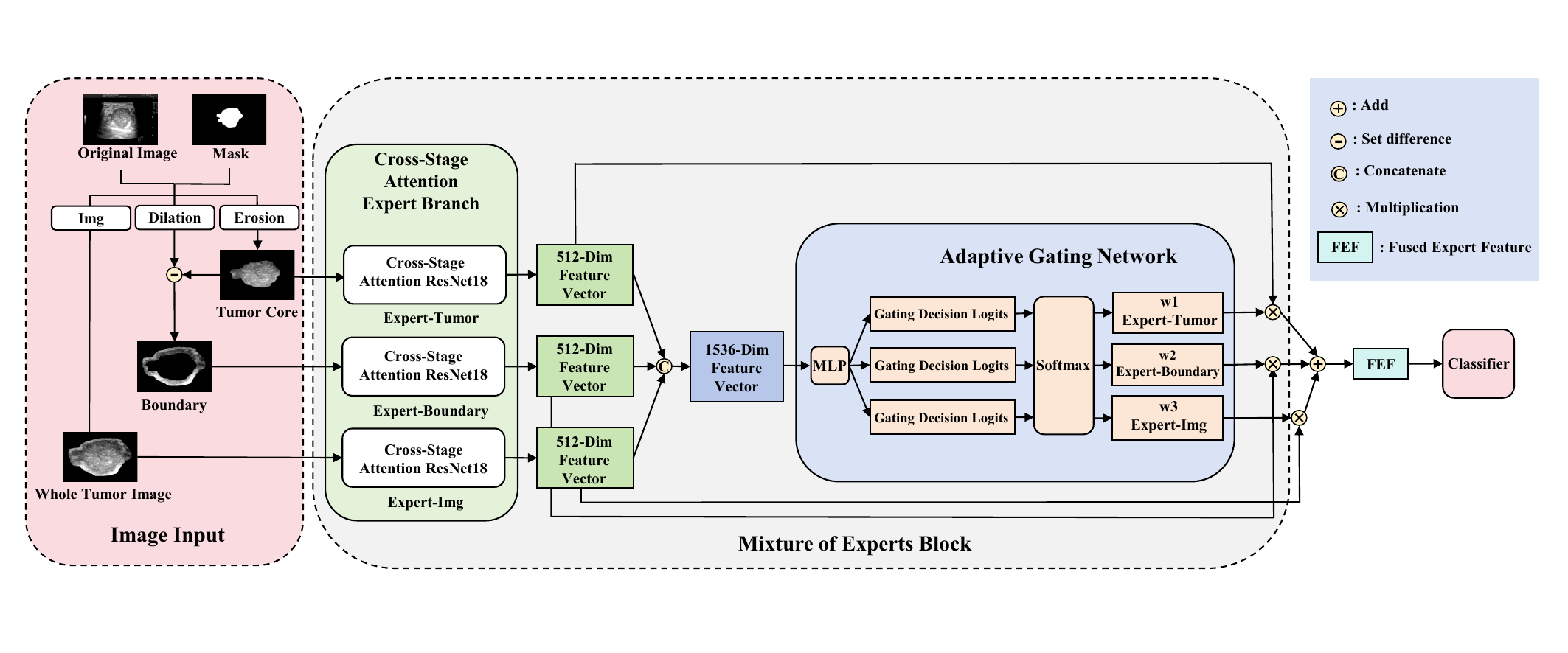}
\caption {Overall architecture of the proposed CSA-MoE-Net. It comprises a Mixture of Experts Block with three dedicated expert branches, a Cross-Stage Attention ResNet-18 backbone, and an Adaptive Gating Network for feature fusion and classification. The final fused feature is denoted as Fused Expert Feature (FEF).}
\label{fig:architecture}
\end{figure*}

\section{Methodology}
The breast ultrasound classification model proposed in this paper adopts Cross-Stage Attention ResNet-18 as the backbone network and constructs a clinically driven three-branch MoE Block. The overall model follows the technical route of “Multi-View Feature extraction — Cross-Stage Attention enhancement — Dynamic Gating Fusion — Classification Decision." Based on the clinical assessment logic of radiologists for  “Whole Tumor Image – Tumor Core – Boundary assessment," three complementary expert branches are designed: Expert-Img, Expert-Tumor, and Expert-Boundary, which independently extract multidimensional features. Each branch fuses multi-level semantic information through the Cross-Stage Attention mechanism to strengthen the expression of key features. Finally, an Adaptive Gating Network dynamically weights and fuses the outputs of the three branches, enabling comprehensive capture of lesion morphology, texture, and contextual information. This architecture deeply integrates clinical diagnostic priors with deep learning techniques, effectively enhancing feature discriminability and classification accuracy. The overall model architecture is shown in Fig.1.

\subsection{Image Preprocessing and Multi-view Data Generation}

Due to inherent issues in breast ultrasound images, including speckle noise, uneven grayscale distribution, and large variations in lesion size, standardized preprocessing is necessary before model training to ensure data consistency and enhance model robustness [22]. This paper standardizes and constructs multi-view samples based on lesion masks, with the overall process shown in the “Image Input" section of Fig.1.

First, image and mask registration are performed. Original grayscale ultrasound images are converted to three-channel pseudo-RGB format and resized to 224×224 pixels. Manually annotated lesion masks are adjusted to match image resolution using nearest-neighbor interpolation, ensuring precise spatial correspondence for subsequent segmentation [23].

Next, Tumor Core images are constructed. Using registered masks as constraints, five morphological erosion operations with a 3×3 structuring element remove boundary noise, retaining the solid lesion body. Pixel-wise multiplication of the mask with the original image extracts only the tumor region, yielding a core image focused on internal texture and echo features.

Finally, Boundary feature regions are extracted. The original mask undergoes five morphological dilation operations. The difference between the dilated and original masks yields a narrow boundary band, which filters the original image to generate a boundary region image containing only contour, spicule, and margin information.

This process generates three input samples: the Whole Tumor Image, Tumor Core, and Boundary, eliminating manual cropping and ROI extraction and solving issues of cumbersome workflows and poor reproducibility [12]. During training, random horizontal flipping, vertical flipping, and ±15° rotation are applied for data augmentation, expanding sample distribution, and alleviating overfitting without altering pathological features [24].

\subsection{The Overall Logic of The MoE Block}
The three-branch MoE Block in this study is essentially a “divide-and-conquer---fusion" feature learning system. It decomposes breast ultrasound classification into three sub-tasks handled by specialized experts, whose outputs are integrated via an Adaptive Gating Network [16]. Compared with single-branch networks, MoE enables each expert to focus on feature modeling from a specific perspective, extracting deeper discriminative information, while the dynamic gating mechanism adaptively allocates computational resources based on input characteristics, improving efficiency and generalization.
Unlike existing medical imaging MoE methods, the expert division in this work aligns closely with the clinical reading workflow of physicians, rather than abstract feature types. Expert-Img learns lesion context; Expert-Tumor extracts internal echogenicity and microcalcifications; and Expert-Boundary captures margin irregularity and spiculations. This anatomy-based division aligns the decision of the model process with clinical cognition, improving interpretability [25]. Experiments show the architecture significantly enhances recognition of complex lesions, with outstanding performance on blurred-boundary and tiny lesions.

\subsection{Expert Network Architecture}
\subsubsection{Expert Network Infrastructure——ResNet-18}
All three expert branches share the same underlying network structure, adopting Cross-Stage Attention ResNet18 as the backbone. During training, each branch optimizes independently. ResNet-18 is chosen for its inherent residual connections, which solve gradient vanishing and degradation problems. Its moderate depth and 11.2M parameters also offer good generalization on small medical datasets, avoiding overfitting in deeper networks [26].ResNet-18 includes a Stem layer, four Residual Stages, Global Average Pooling (GAP), and a Fully Connected layer. The Stem converts [B,3,224,224] inputs to [B,64,56,56] via convolution, batch normalization, ReLU, and max pooling. Each residual stage contains two residual blocks with “main path + shortcut path" structures: the main path learns residuals via two 3×3 convolutions, while the shortcut uses identity mapping or 1×1 convolution for dimension matching.
Residual block forward propagation is formalized as:
\begin{equation}
\boldsymbol{y} = \mathcal{F}(\boldsymbol{x}, \{W_i\}) + \boldsymbol{x}
\end{equation}

\begin{equation}
\boldsymbol{y} = \mathcal{F}(\boldsymbol{x}, \{W_i\}) + \boldsymbol{W}_s \cdot \boldsymbol{x}
\end{equation}

\begin{equation}
\mathcal{F}(\boldsymbol{x}, \{W_i\}) = W_2 \cdot \sigma\big(\text{BN}(W_1 \cdot \boldsymbol{x})\big)
\end{equation}

\noindent
where $\boldsymbol{x}$ denotes the input feature tensor of the residual block;
$\boldsymbol{y}$ denotes the output feature tensor of the residual block;
$\mathcal{F}(\boldsymbol{x},\{W_i\})$ represents the residual mapping to be learned in the main path;
$\{W_i\}$ is the weight set of convolutional layers in the main path, i.e., $\{W_1,W_2\}$;
$W_1$ and $W_2$ are the weight matrices of two consecutive convolutional layers;
$\sigma(\cdot)$ denotes the ReLU activation function;
$\text{BN}(\cdot)$ represents the batch normalization operation;
$\boldsymbol{W}_s$ is the weight matrix of $1\times1$ convolution used for dimension matching in the shortcut branch. 

As stages deepen, channels increase from 64 to 512, and spatial size is downsampled, enabling abstraction from low-level to high-level features. GAP aggregates $[B, 512, 7, 7]$ feature maps into 512-dimensional vectors, reducing parameters and overfitting risk [4].

\subsubsection{Cross-Stage Attention Module}
Traditional channel attention mechanisms (e.g., SE-Net) only perform channel recalibration within a single layer, relying on Global Pooling that may discard critical local details such as lesion boundaries and internal echoes. Moreover, they lack interaction across hierarchical features, making it hard to fuse shallow details and deep semantics [17]. To solve these problems, this paper designs a Cross-Stage Attention module for ResNet-18, which builds dynamic connections among the four residual stages to realize multi-scale feature fusion and adaptive channel recalibration. The module includes three steps. 

First, feature aggregation and dimension mapping. For the feature maps $F_1$–$F_4$ ($[B,64,56,56]$, $[B,128,28,28]$, $[B,256,14,14]$, $[B,512,7,7]$) output by the four residual stages of ResNet-18, adaptive average pooling is applied to each layer to compress the features into Whole Tumor Image vectors with the same number of channels; subsequently, through 8 independent linear mapping layers (Key and Value branches each with 4 layers), the feature vectors of different dimensions are uniformly mapped to 128 dimensions, eliminating the dimension differences between layers and laying the foundation for cross-layer interaction.

Second, Cross-Stage Attention weight computation: A learnable query vector is used to calculate similarity with stage-wise key vectors. Softmax normalization yields layer attention weights $W_1$–$W_4$, which reflect the contribution of each stage to classification.

Third, high-level feature recalibration: The weighted 128-dimensional feature is processed by an MLP, compressed to 64 dimensions, and restored to 512 dimensions with Sigmoid activation to produce channel weights. These weights are multiplied by the final stage feature $F_4$ to enhance informative channels and suppress irrelevant ones.

The above cross-stage attention mechanism can be formally described as follows. 
Let the output feature map of the $k$-th residual stage be 
$\boldsymbol{F}_k \in \mathbb{R}^{B\times C_k \times H_k \times W_k}~(k = 1,2,3,4)$, 
where $C_k \in \{64, 128, 256, 512\}$.

\textbf{Phase 1 — Feature Aggregation and Dimension Mapping}
\begin{equation}
\boldsymbol{z}_k = \text{AAP}(\boldsymbol{F}_k) = \frac{1}{H_k W_k} \sum_{i,j} \boldsymbol{F}_k(:,:,i,j)
\end{equation}

\begin{equation}
\boldsymbol{K}_k = \boldsymbol{W}_k^K \cdot \boldsymbol{z}_k,\quad 
\boldsymbol{V}_k = \boldsymbol{W}_k^V \cdot \boldsymbol{z}_k
\end{equation}
where $\text{AAP}(\cdot)$ denotes adaptive average pooling; 
$\boldsymbol{z}_k \in \mathbb{R}^{C_k}$ is the pooled feature vector; 
$\boldsymbol{W}_k^K$ and $\boldsymbol{W}_k^V$ are the key and value projection matrices, 
which map features into a unified 128-dimensional space.

\textbf{Phase 2 — Cross-Stage Attention Weight Calculation}
\begin{equation}
\alpha_k = \text{softmax}_k\left( \frac{\boldsymbol{Q} \cdot \boldsymbol{K}_k^\mathrm{T}}{\sqrt{128}} \right)
\end{equation}
where $\boldsymbol{Q} \in \mathbb{R}^{1\times 128}$ is a learnable query vector, 
$\alpha_k$ denotes the normalized attention weight of the $k$-th stage, satisfying $\sum_{k=1}^4 \alpha_k = 1$.

\textbf{Phase 3 — Channel-wise Feature Recalibration}
\begin{equation}
\boldsymbol{F}_\text{fused} = \sum_{k=1}^4 \alpha_k \cdot \boldsymbol{V}_k
\end{equation}
 
\begin{equation}
\boldsymbol{W}_\text{ch} = \sigma\Big( \boldsymbol{W}_\text{up} \cdot \text{ReLU}\big( \boldsymbol{W}_\text{down} \cdot \boldsymbol{F}_\text{fused} \big) \Big)
\end{equation}

\begin{equation}
\hat{\boldsymbol{F}}_4 = \boldsymbol{W}_\text{ch} \odot \boldsymbol{F}_4
\end{equation}
where $\boldsymbol{F}_\text{fused}$ represents the weighted fused feature; 
$\boldsymbol{W}_\text{down}$ and $\boldsymbol{W}_\text{up}$ constitute the bottleneck MLP; 
$\sigma(\cdot)$ denotes the Sigmoid activation function; 
$\boldsymbol{W}_\text{ch}$ is the channel-wise attention weight; 
$\odot$ indicates the channel-wise element-wise multiplication operation.

Compared with existing Cross-Stage Attention methods, this module achieves dynamic assessment of layer importance via a learnable query vector instead of fixed fusion weights, making it more adaptable to heterogeneous features of breast ultrasound lesions [27]. It brings little extra computation but significantly improves the ability to locate critical regions, especially for blurred boundaries and tiny lesions. The Cross-Stage Attention architecture is shown in Fig.2.

\subsection{Multi-Expert Fusion Adaptive Gating Network and Model Decision End}
To achieve adaptive weighted fusion of the three expert branch features, we propose a two-layer MLP gating network guided by clinical diagnostic logic. Unlike static weighting schemes, this network dynamically modulates the contribution of each expert based on input lesion characteristics: it emphasizes Expert-Img and Expert-Tumor for well-defined benign lesions, while automatically upweighting Expert-Boundary for malignant lesions with irregular margins [28].

Let $F_I, F_T, F_B \in \mathbb{R}^{512}$ denote the outputs of the three experts. The gating network takes the concatenated feature $F_{\text{concat}} = [F_I; F_T; F_B] \in \mathbb{R}^{1536}$ as input, and computes the expert weights via:
\begin{equation}
w = \text{Softmax}\left( W_2 \cdot \text{ReLU}\left( \text{BN}\left( W_1 F_{\text{concat}} + b_1 \right) \right) + b_2 \right)
\end{equation}
where $W_1 \in \mathbb{R}^{64 \times 1536}$, $W_2 \in \mathbb{R}^{3 \times 64}$ are learnable weights, and dropout (rate=0.2) is applied after the first layer for regularization.

The final fused feature is obtained by weighted summation:
\begin{equation}\label{eq:fused}
F_{\text{FEF}} = w_1 F_I + w_2 F_T + w_3 F_B \in \mathbb{R}^{512}
\end{equation}
This fusion strategy mimics the diagnostic reasoning of radiologists by integrating complementary Whole Tumor Image, Tumor Core, and Boundary information. The 512-dimensional fused feature is then passed to a fully connected layer with Sigmoid activation for binary classification.

\begin{figure*}[!t]
    \centering
    \includegraphics[width=1\linewidth]{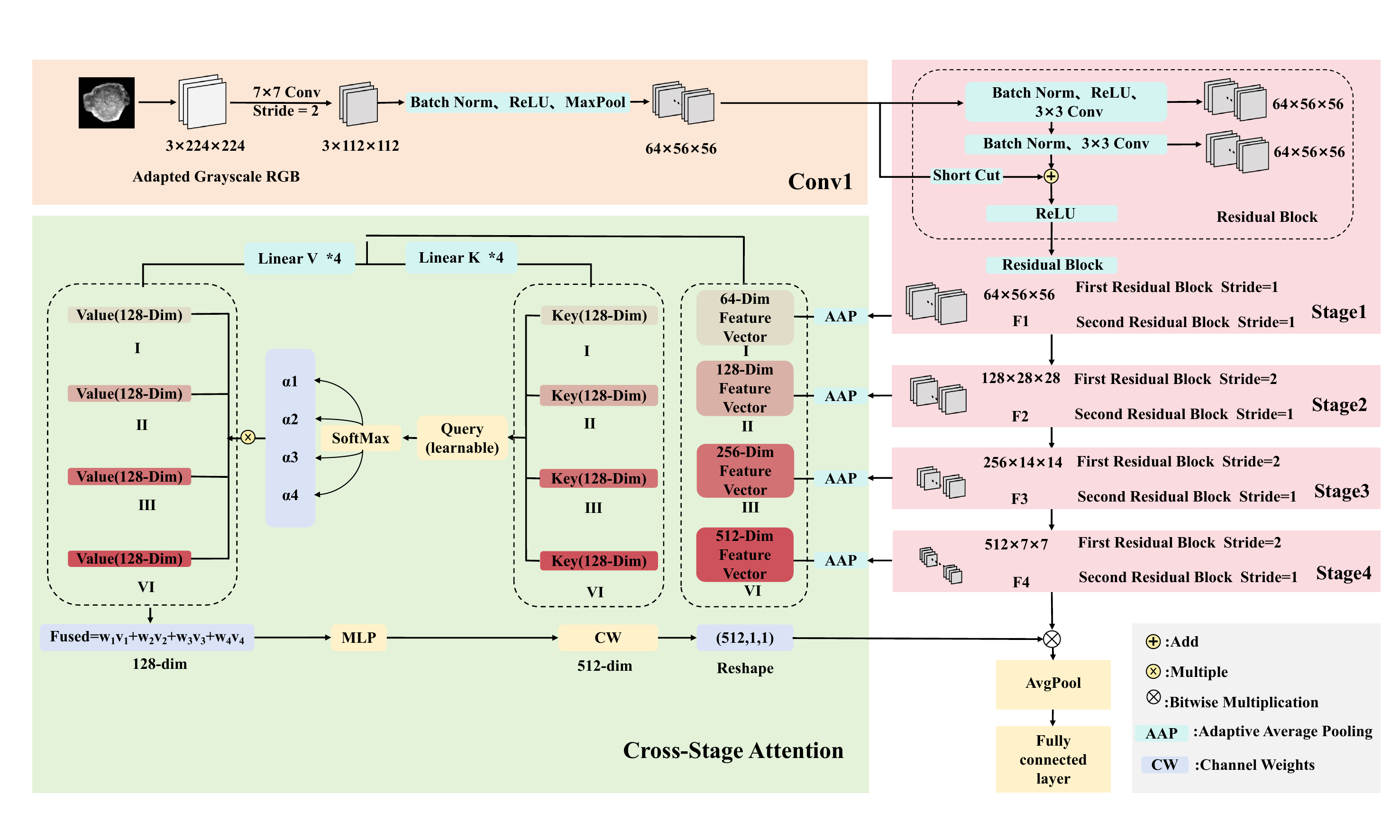}
    \caption{Overall Architecture of Cross-Stage Attention Mixture of Experts Network for Breast Cancer Ultrasound Nodule Diagnosis}
    \label{fig:placeholder}
\end{figure*}

\section{Dataset Configuration and Experimental Results}
\subsection{Dataset Configuration and Experimental setup}
\subsubsection{Dataset Configuration}
This experiment uses real clinical breast ultrasound images to evaluate the overall performance of the proposed CSA-MoE-Net. All images were provided by partner hospitals, with proper approvals obtained. Patient privacy was protected by anonymizing all data, retaining only valid lesion regions.

The dataset contains 2,129 images (1,108 benign and 1,021 malignant), each with a corresponding lesion mask annotated by senior radiologists, providing a reliable basis for multi-view region extraction and feature analysis [14].
A stratified sampling approach was used for dataset splitting: 15\% of the data was set aside as the test set, and 17.65\% of the remaining 85\% was used as the validation set. The final split was approximately 70\% training, 15\% validation, and 15\% test. Random seed 42 was fixed throughout to ensure reproducibility of dataset division and experimental results.

\subsubsection{Experimental setup}
All experiments in this study were developed and implemented using the PyTorch deep learning framework. The backbone network, ResNet-18, was initialized with public ImageNet1K-V1 pre-trained weights for transfer learning [5]. Experiments were run with GPU acceleration on an NVIDIA GeForce RTX 4060 Laptop GPU.

The Adam optimizer was used for model training, with an initial learning rate of 5e-5 and no additional weight decay. Cross-entropy loss was applied for the binary classification task. Training hyperparameters were unified: 30 epochs, batch size of 32, and 4 worker threads for asynchronous data loading. Training samples were randomly shuffled, while validation and test sets were fed to avoid evaluation interference.

A ReduceLROnPlateau strategy was used for adaptive learning rate adjustment, reducing the rate by half if validation loss did not decrease for 3 consecutive epochs, with a minimum learning rate of 1e-6 to prevent stagnation. Early stopping was applied by saving the model with the best validation accuracy during training and evaluating it on the test set after training. To eliminate random errors, all random seeds (random, numpy, torch, torch.cuda) were set to 42, ensuring stable and reproducible results.

\subsection{Analysis of Experimental Results}
\subsubsection{Analysis of CSA-MoE-Net Results}
This section presents a comprehensive performance evaluation of the proposed CSA-MoE-Net. Evaluation metrics include Accuracy, Precision, Recall, F1-score, and Area Under the Curve (AUC). The experiment was independently repeated 20 times, and the average and standard deviation (SD) of the test set results were reported to eliminate the impact of random fluctuations from a single run. The overall experimental results of the model are shown in Table 1.

\begin{table}[!ht]
\centering
\renewcommand{\arraystretch}{2} 
\caption{Experimental results based on the CSA-MoE-Net }
\label{tab:result_overall}
\begin{tabular}{lccccc}
\toprule
Data & Accuracy(\%) & Precision(\%) & Recall(\%) & F1(\%) & AUC(\%) \\
\midrule
Average & 96.33 & 94.09 & 98.53 & 96.25 & 99.50 \\
SD & $\pm$0.7 & $\pm$1.2 & $\pm$0.7 & $\pm$0.7 & $\pm$0.1 \\
\bottomrule
\end{tabular}
\end{table}

As shown in Table 1, CSA-MoE-Net achieves an AUC of 99.50\%, indicating strong discriminative ability between benign and malignant lesions. The recall rate reaches 98.53\%, demonstrating high sensitivity for identifying malignant lesions and effectively reducing the risk of missed diagnoses in clinical practice. In breast cancer screening, the cost and life impact of missed diagnoses far outweigh those of unnecessary examinations caused by false positives, making high recall a core design goal for clinical decision support systems. The model’s precision of 94.09\% and F1-score of 96.25\% indicate a low false-positive rate while minimizing missed cases. The standard deviation across all evaluation metrics is less than 2\%, indicating stable outputs, low fluctuation, and strong generalization across repeated experiments. This convergence behavior aligns with the training characteristics of medical small-sample datasets, where the model can learn key discriminative features after only a few iterations, while prolonged training may lead to overfitting [29].

\begin{figure}[!ht]
    \centering
    \includegraphics[width=1\linewidth]{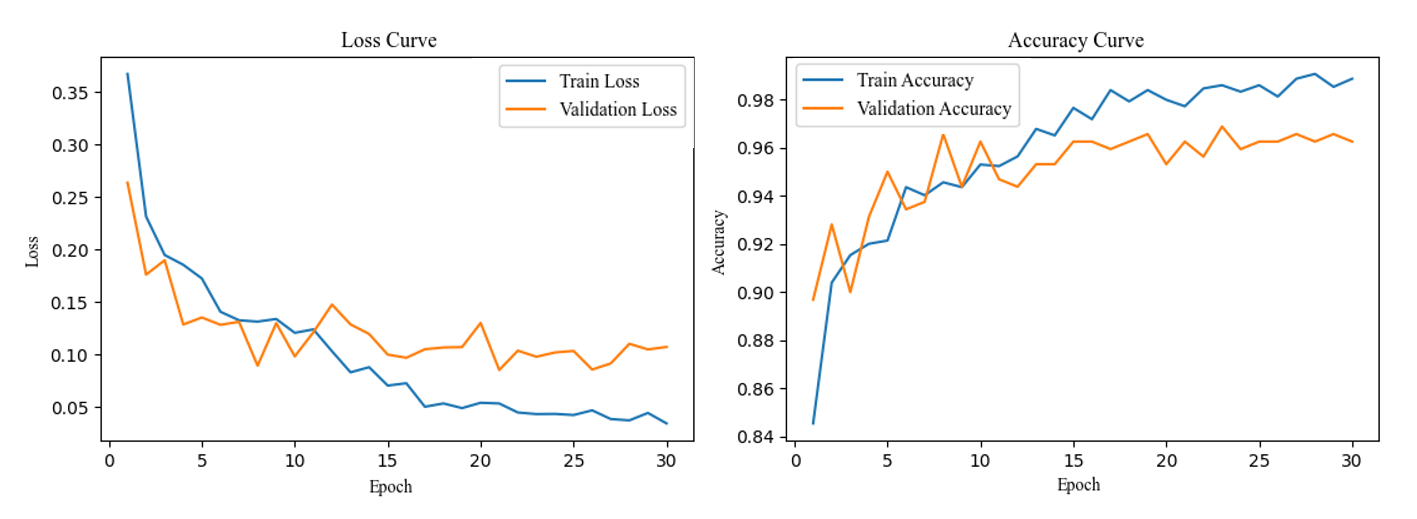}
    \caption{The curves showing the changes of Loss and Accuracy in the training set and validation set of the CSA-MoE-Net over epochs}
    \label{fig:placeholder}
\end{figure}

Fig.3 shows the curves for loss and accuracy of the CSA-MoE-Net on the training and validation sets over epochs. Training loss drops rapidly during the first 5 epochs and then stabilizes, while training accuracy plateaus after 10 epochs. Validation loss and accuracy converge synchronously, with the final gap between training and validation accuracy being less than 2\%, and no obvious overfitting is observed, which confirms the model’s good generalization ability.

To visually demonstrate the adaptive fusion mechanism of the three-expert MoE architecture during training, Fig.4 shows the changes in gating weights for each expert network over epochs. Solid lines represent training set weights, dashed lines represent validation set weights, and blue, yellow, and green correspond to the Expert-Tumor, Expert-Boundary, and Expert-Img, respectively.

\begin{figure}[!ht]
    \centering
    \includegraphics[width=1\linewidth]{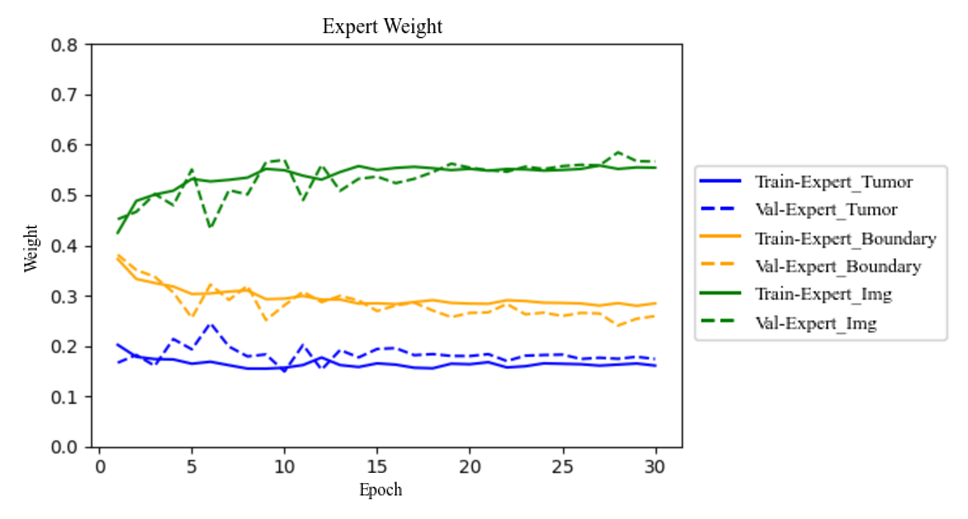}
    \caption{Curves showing the changes of expert weights in the training set and validation set of the CSA-MoE-Net over different epochs}
    \label{fig:placeholder}
\end{figure}

Overall, the  Expert-Img quickly becomes dominant, with its weight rising to about 0.55 early in training and remaining stable, indicating the model’s reliance on global context for initial decisions. The weight of Expert-Boundary starts around 0.35, decreases to about 0.25 within the first 10 epochs, and then stabilizes. The weight of Expert-Tumor is relatively low and stable, around 0.18. Validation set weights closely match training set weights, showing good generalization and no significant overfitting. 

After five epochs, the weights settle into a steady ratio of roughly “Expert-Img : Expert-Boundary: Expert-Tumor $\approx$ 3: 1.4: 1". This fast convergence indicates that the MoE gating mechanism quickly learns the complementary relationships among experts, allowing the model to achieve stable fusion without prolonged exploration [30].

\begin{table*}[!t]
\centering
\renewcommand{\arraystretch}{2.5}
\caption{Comparison of ablation experiment results of CSA-MoE-Net, ResNet-MoE and ResNet-18 baseline models}
\label{tab:ablation_main}
\begin{tabularx}{\linewidth}{@{}l *{7}{>{\centering\arraybackslash}X}@{}}
\toprule
Model & \makecell{Cross-Stage\\Attention} & \makecell{Mixture of\\Experts} & Accuracy(\%) & Precision(\%) & Recall(\%) & F1(\%) & AUC(\%) \\
\midrule
CSA-MoE-Net & $\surd$ & $\surd$ & 96.33$\pm$0.70 & 94.09$\pm$1.20 & 98.53$\pm$0.70 & 96.25$\pm$0.70 & 99.50$\pm$0.10 \\
ResNet-MoE & $\times$ & $\surd$ & 94.73$\pm$0.50 & 94.11$\pm$1.10 & 94.94$\pm$1.40 & 94.51$\pm$0.50 & 99.19$\pm$0.20 \\
ResNet-18 & $\times$ & $\times$ & 93.32$\pm$0.30 & 93.39$\pm$0.20 & 93.16$\pm$0.80 & 93.27$\pm$0.40 & 94.08$\pm$1.70 \\
\bottomrule
\end{tabularx}
\end{table*}

\subsubsection{Ablation Experiment}
An ablation study was conducted by experimenting with both the ResNet-18-based Mixture of Experts network model (ResNet-MoE) and the ResNet-18 baseline model. All experiments used the same dataset splits and training hyperparameters, and were independently repeated 20 times to calculate the average and standard deviation of the test set results. The performance metrics of CSA-MoE-Net, ResNet-MoE, and ResNet-18 were summarized in a single table for comparison, as shown in Table 2. (All metric values in the table are presented as percentages.)

Comparing CSA-MoE-Net with ResNet-MoE, removing the Cross-Stage Attention mechanism led to a drop in accuracy from 96.33\% to 94.73\% (down 1.60\%), recall from 98.53\% to 94.94\% (down 3.59\%), F1-score from 96.25\% to 94.51\%, and AUC from 99.50\% to 99.19\%. Precision remained nearly unchanged, indicating that Cross-Stage Attention mainly helps the model better localize key lesion regions and reduce false negatives, rather than simply improving positive prediction precision. These results demonstrate the significant role of Cross-Stage Attention in guiding the model to focus on discriminative regions in ultrasound images.

Additionally, according to Fig.3 and Fig.5, the loss and accuracy curves for ResNet-MoE are similar to those for CSA-MoE-Net in terms of trends, convergence speed, and overfitting, suggesting that Cross-Stage Attention does not alter training dynamics or gating strategy, but mainly refines feature representation and enhances deep feature discrimination. This leads to the conclusion that Cross-Stage Attention notably improves recall, but has minimal effect on precision, highlighting its role in boosting lesion localization sensitivity without changing the classification threshold.

\begin{figure}[!ht]
    \centering
    \includegraphics[width=1\linewidth]{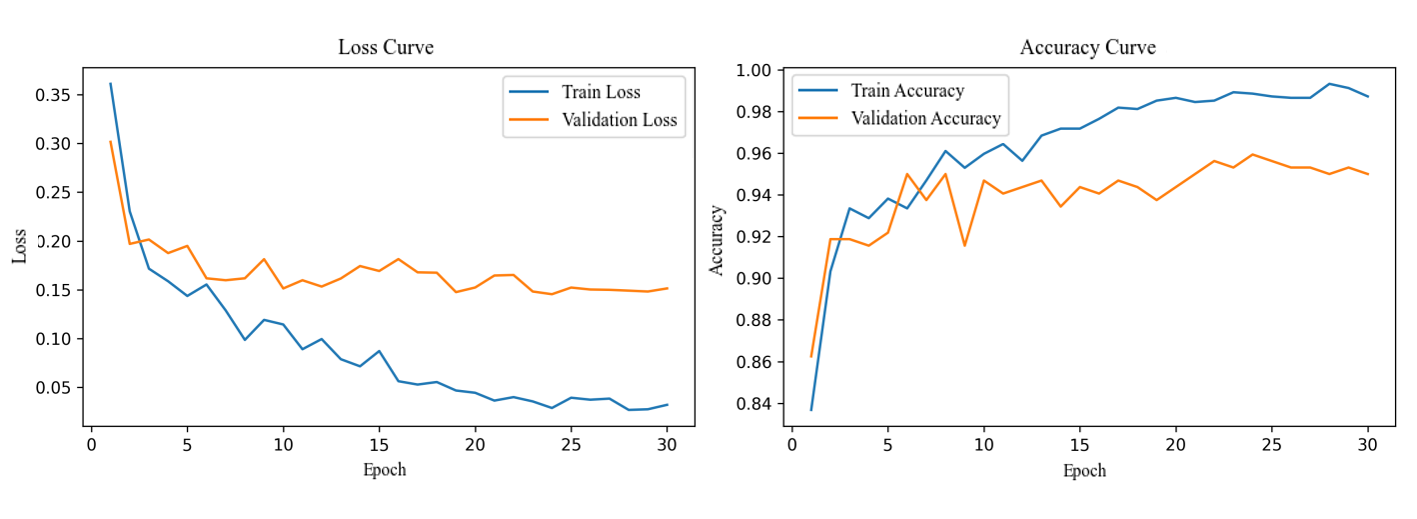}
    \caption{The curves of Loss and Accuracy of the ResNet-MoE model training set and validation set over the Epochs}
    \label{fig:placeholder}
\end{figure}

The weight curves of ResNet-MoE and CSA-MoE-Net are plotted in Fig.6 for a visual comparison.Fig.6 shows a highly consistent pattern: the Expert-Img always dominates, with its weight quickly rising and stabilizing at the top, while the Expert-Boundary and Expert-Tumor remain as secondary contributors, forming a “one main, two auxiliary” structure. This pattern remains stable regardless of the inclusion of Cross-Stage Attention, indicating that the gating network’s top-level assessment of expert importance is unchanged.

\begin{figure}[!ht]
    \centering
    \includegraphics[width=1\linewidth]{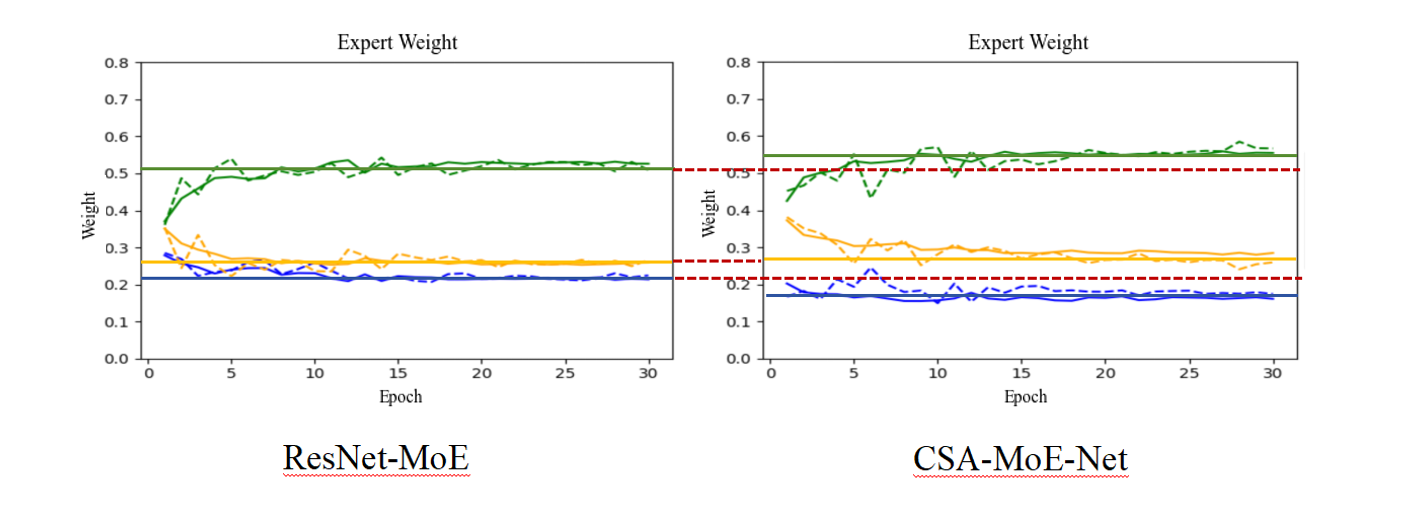}
    \caption{Comparison of Weights between ResNet-MoE Model and CSA-MoE-Net}
    \label{fig:placeholder}
\end{figure}

The only minor differences are a slight decrease in the Expert-Img’s weight and a slight increase in the Expert-Tumor’s weight when Cross-Stage Attention is removed, with the Expert-Boundary’s weight nearly unchanged. This shift suggests that the gating network tries to compensate for the loss in deep feature discrimination by increasing reliance on the Expert-Tumor. However, ablation results show this compensation is insufficient—recall still drops by 3.59\% and AUC by 0.31\%—highlighting that Cross-Stage Attention’s core value lies in enhancing the discriminative power of the Expert-Img, enabling better performance without disrupting the overall expert weighting structure.

\begin{table*}[!t]
\centering
\renewcommand{\arraystretch}{2.5}
\caption{Ablation Results of Different Expert Branches}
\label{tab:ablation_three_expert}
\begin{tabularx}{\linewidth}{@{}X *{5}{>{\centering\arraybackslash}X}@{}}
\toprule
Model & Accuracy(\%) & Precision(\%) & Recall(\%) & F1(\%) & AUC(\%) \\
\midrule
Expert-Img+Boundary & 95.94$\pm$0.50 & 93.98$\pm$0.70 & 98.04$\pm$0.60 & 95.96$\pm$0.50 & 99.54$\pm$0.10 \\
Expert-Img+Tumor    & 95.27$\pm$0.53 & 93.35$\pm$1.45 & 97.08$\pm$0.70 & 95.13$\pm$0.58 & 99.49$\pm$0.19 \\
Full 3-Expert       & 96.33$\pm$0.70 & 94.09$\pm$1.20 & 98.53$\pm$0.70 & 96.25$\pm$0.70 & 99.50$\pm$0.10 \\
\bottomrule
\end{tabularx}
\end{table*}

Further comparison of ResNet-MoE and the baseline ResNet-18 shows that removing the MoE module leads to a clear drop in performance. As shown in Table 2, accuracy decreases from 94.73\% to 93.32\%, precision from 94.11\% to 93.39\%, recall from 94.94\% to 93.16\%, F1-score from 94.51\% to 93.27\%, and AUC drops most sharply from 99.19\% to 94.08\% (down by 5.11\%). The baseline ResNet-18 also has a much higher AUC standard deviation (±1.70\%), indicating that the MoE architecture not only improves classification but also enhances robustness to data variations [31]. By leveraging multi-expert collaboration and gating, MoE enables adaptive feature selection and better handles complex ultrasound patterns.

\begin{figure}[!ht]
    \centering
    \includegraphics[width=1.1\linewidth]{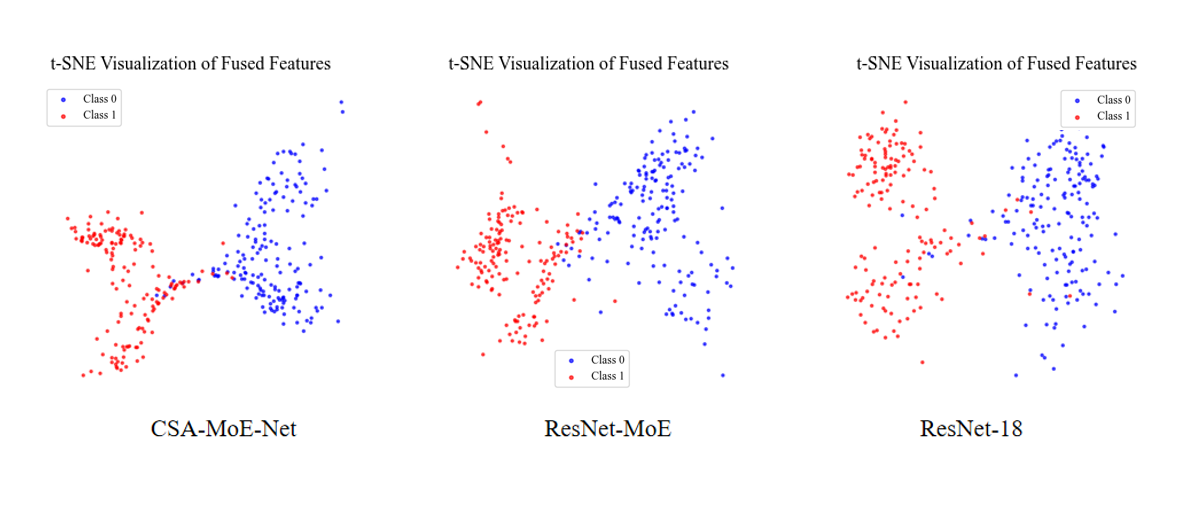}
    \caption{The t-SNE diagrams of the three models tested in the ablation experiment}
    \label{fig:placeholder}
\end{figure}

Fig.7 shows t-SNE feature visualizations for CSA-MoE-Net, ResNet-MoE, and baseline ResNet-18. Feature visualization intuitively reflects the model’s learned feature distribution and class separability [32]. The full CSA-MoE-Net displays the clearest class boundaries, with benign (blue) and malignant (red) samples forming tight, well-separated clusters. Removing Cross-Stage Attention increases boundary mixing, with more benign samples leaking into malignant clusters, indicating reduced discrimination, especially for hard-to-classify cases. The baseline ResNet-18 model shows the worst clustering, with significant overlap between classes and loose distribution, meaning features lack clear separation after dimensionality reduction. These results confirm that Cross-Stage Attention enhances deep feature consistency, while the MoE structure enriches feature diversity—together, they bring similar samples closer and push dissimilar samples farther apart in feature space.

In summary, the ablation experiments show that Cross-Stage Attention and the MoE architecture are complementary in improving model performance. Cross-Stage Attention focuses on selecting important features across network layers, enhancing the semantic discriminability of single expert networks by adaptively recalibrating channel weights. In contrast, the MoE architecture emphasizes adaptive fusion of multi-region features, dynamically integrating local and global information through gating among the three expert branches. Together, they enable the model to achieve optimal or near-optimal results across all metrics.

\subsubsection{Validation of the effectiveness of multi-regional perspectives}
To assess the contribution of each expert network within the MoE framework, we conducted comparative experiments by removing one expert(All retain Expert-Img) at a time while keeping the overall architecture unchanged. All experiments used consistent dataset splits and training parameters, with results averaged over 20 runs and reported as mean and standard deviation on the test set. Results are shown in Table 3 (all metrics are percentages).

Comparing CSA-MoE-Net with Expert-Img+Boundary (removing the Expert-Tumor), accuracy, recall, and F1-score all drop slightly, though AUC remains similar. This indicates that Boundary and Whole Tumor Image information capture most discriminative features, but losing internal texture details still impacts performance, especially recall.

Comparing CSA-MoE-Net with Expert-Img+Tumor (removing the Expert-Boundary), the performance drops more noticeably—accuracy, precision, recall, F1, and AUC all decrease, with recall down 1.45\% from the full model. This suggests that tumor boundary features (such as irregular boundaries and spiculations) are more critical for classification than internal features, aligning with clinical experience that boundary characteristics are key malignancy indicators.

\begin{figure}[!ht]
    \centering
    \includegraphics[width=1\linewidth]{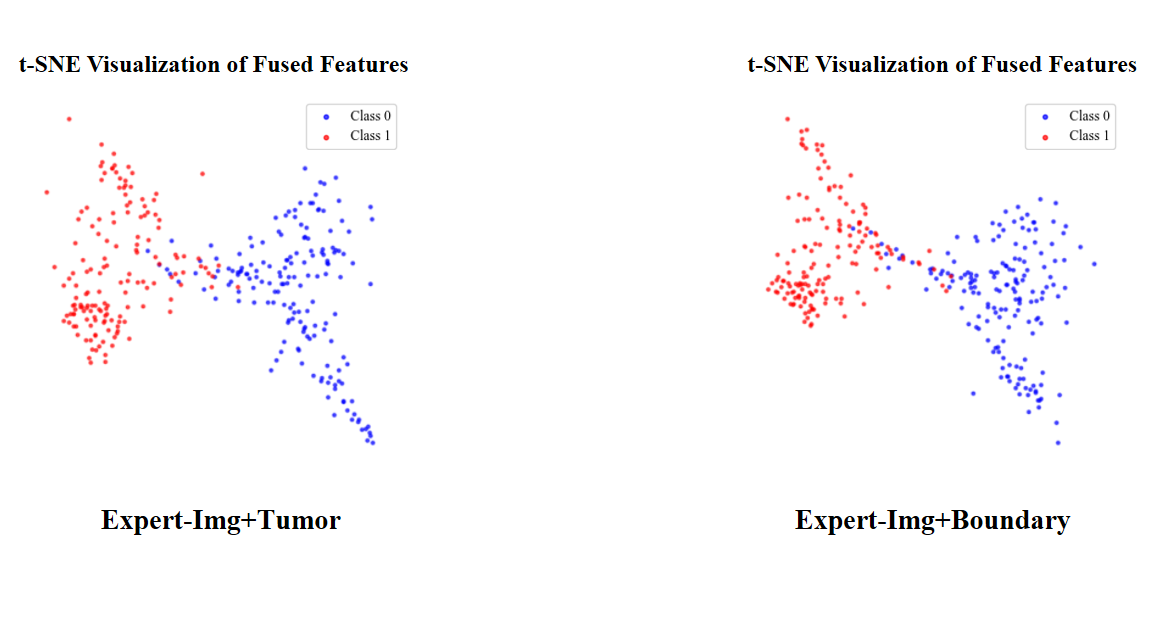}
    \caption{The t-SNE diagrams of the Expert-Img+Boundary and Expert-Img+Tumor models}
    \label{fig:placeholder}
\end{figure}

Fig.8 shows t-SNE visualizations for both configurations: Expert-Img+Boundary achieves tighter, more separated clusters, while Expert-Img+Tumor shows greater overlap between classes and a more loosely distributed pattern. This further confirms that the Expert-Boundary plays a crucial role in extracting discriminative features and improving class separability.

\begin{table*}[!t]
  \centering
  \caption{Performance parameters, model parameter quantities and FLOPs results of multiple models}
  \label{tab:combined_results}
  \setlength{\tabcolsep}{5pt}
  \renewcommand{\arraystretch}{2}
  \begin{tabular*}{\linewidth}{@{\extracolsep{\fill}} l c c c c c c c}
    \hline
    Model & Acc(\%) & Pre(\%) & Recall(\%) & F1(\%) & AUC(\%) & Params(M) & FLOPs(G) \\
    \hline
    ResNet-18          & 93.32 & 93.39 & 93.16 & 93.27 & 94.08 & 11.178  & 1.824  \\
    ResNet-MoE         & 94.73 & 94.11 & 94.94 & 94.51 & 99.19 & 33.958  & 5.471  \\
    \textbf{CSA-MoE-Net } & \textbf{96.33} & \textbf{94.09} & \textbf{98.53} & \textbf{96.25} & \textbf{99.50} & 34.820  & 5.472  \\
    \hline
    DenseNet-121       & 94.18 & 95.42 & 92.29 & 93.82 & 98.75 & 6.956   & 2.896  \\
    DenseNet-MoE       & 95.00 & 95.59 & 93.90 & 94.73 & 99.13 & 22.720  & 9.012  \\
    \textbf{CSA-MoE-DenseNet} & \textbf{95.78} & \textbf{95.83} & \textbf{95.34} & \textbf{95.57} & \textbf{99.23} & 23.310  & 9.013  \\
    \hline
    VGG-16             & 94.22 & 94.39 & 93.47 & 93.93 & 99.06 & 134.269 & 15.466 \\
    VGG-MoE            & 95.19 & 92.80 & 97.52 & 95.09 & 99.40 & 415.501 & 46.040 \\
    \textbf{CSA-MoE-VGG}   & \textbf{96.88} & \textbf{97.99} & \textbf{95.42} & \textbf{96.69} & \textbf{99.14} & 416.412 & 46.041 \\
    \hline
  \end{tabular*}
\end{table*}

Comparing CSA-MoE-Net with Expert-Img+Boundary (without the Expert-Tumor), there is a slight drop in accuracy, recall, and F1-score, while AUC remains about the same. This indicates that Boundary and Whole Tumor Image information capture most discriminative features, but the loss of internal texture still affects performance, especially recall.

\subsubsection{Verification of other CNN models}
The above experiments demonstrate the complementarity of the three-expert architecture. To further verify the generalizability and clinical potential of the proposed modules, we seamlessly integrated Cross-Stage Attention and the MoE architecture into other mainstream backbones (VGG-16, DenseNet-121) for comparison. Both configurations retained the Cross-Stage Attention (with only channel dimension adaptation) and three-expert MoE structure, only changing the feature extractor. Similar ablation experiments were performed, with each experiment repeated 20 times and results reported as mean and standard deviation on the test set to reduce random fluctuations. Results are shown in Table 4. (The meanings of "Acc" and "Pre" are "Accuracy" and "precision" respectively.All metrics are percentages)

Ablation results in Table 4 show a clear performance gradient for both VGG-16 and DenseNet-121: full model $>$ MoE only$>$baseline. This confirms the effectiveness and synergy of Cross-Stage Attention and the MoE structure. For VGG-16, accuracy rises from 94.22\% (baseline) to 95.19\% (MoE), and further to 96.88\% with both modules; precision also improves, indicating that Cross-Stage Attention helps the model focus on key tumor features. For DenseNet-121, accuracy increases from 94.18\% (baseline) to 95.00\% (MoE), and to 95.78\% with both modules, while AUC also rises notably. This helps address feature redundancy in DenseNet.

VGG-16 achieves the highest accuracy and precision, but has relatively lower recall, suggesting a more conservative approach that could increase false negatives in screening. DenseNet-121’s performance is slightly lower than ResNet-18, but still strong and stable. ResNet-18 achieves the best balance, with top recall (98.53\%) and AUC (99.50\%), thanks to its residual connections, which complement Cross-Stage Attention and the MoE design.

Across all backbones, the proposed modules maintain very low AUC standard deviations, demonstrating robust and stable performance. The results highlight two key advantages: backbone-independence (easy integration into various networks by adapting channel dimensions) and consistent performance gains, with all tested models achieving over 95\% accuracy and 99\% AUC, proving strong generalizability.

Considering the actual usage scenario, the parameters of the model and other related parameters should not be too large. They need to be compatible with the performance of the office computer. Table 4 compares the computational complexity of different baseline models, their MoE variants, and CSA-MoE enhanced versions. Adding the MoE structure roughly triples the parameter count, consistent with the independent branch design: ResNet-MoE has about 3.04 times the parameters of ResNet-18, DenseNet-MoE about 3.27 times DenseNet-121, and VGG-MoE about 3.09 times VGG-16. The Cross-Stage Attention module adds very few extra parameters (about 2–3\%), yet delivers significant performance gains.

All models use full end-to-end training, with no parameter freezing, ensuring maximum feature learning. In summary, the Cross-Stage Attention and MoE modules provide a general enhancement that can boost performance across different backbone networks for breast ultrasound classification, offering a universal optimization solution for medical imaging analysis.

\section{Discussion}
The experimental results fully confirm the performance advantages and generalizability of CSA-MoE-Net, but its design logic, clinical value, and theoretical basis warrant further discussion. The core value of CSA-MoE-Net lies not only in improved prediction accuracy, but also in its explicit alignment with the clinical diagnostic process. Traditional breast ultrasound models often use simplified strategies—such as single-branch global input or manual ROI extraction—which disconnect the model’s decision logic from clinical reasoning and lead to a “black box" with high accuracy but low interpretability.

Moreover, this research offers insights for broader applications: the number and roles of expert networks in MoE can be adjusted for different tasks, and Cross-Stage Attention can be selectively applied to specific layers. Thus, this modular approach is not limited to breast ultrasound, but could also benefit other medical imaging tasks or even other imaging domains.

CSA-MoE-Net replicates the diagnostic workflow of senior radiologists in terms of “Whole Tumor Image – Tumor Core – Boundary analysis" at the structural level through the independent design and dynamic fusion of three Expert-Branch modules: Expert-Tumor (Tumor Core), Expert-Boundary (Boundary), and Expert-Img (Whole Tumor Image). The Expert-Img branch provides spatial context between the lesion and surrounding glands; the Expert-Boundary branch captures malignant indicators such as morphological irregularity and spiculation; the Expert-Tumor branch depicts histological features, including echo homogeneity and microcalcifications. The dynamic weighting mechanism of the Adaptive Gating Network simulates the decision-making behavior of radiologists, who flexibly adjust their focus based on specific lesion manifestations. This structure-as-logic design enables the model to achieve 96.33\% accuracy and 98.53\% recall, while delivering transparent reasoning for “why the model makes such a judgment”. By visualizing gating weights and expert outputs, clinicians can intuitively understand the contribution of features from each view, lowering the clinical adoption barrier of AI-assisted diagnosis systems[36]. In ablation experiments, removing the Expert-Boundary branch decreases recall by 1.45 percentage points and F1-score by 1.12 percentage points, quantitatively verifying the irreplaceability of boundary morphological information in malignant lesion identification—consistent with the clinical consensus that boundary spiculation is a key malignant sign.

The Cross-Stage Attention mechanism designed in this paper breaks the limitation of traditional channel attention (e.g., SE-Net) that only operates on a single layer. By establishing dynamic connections across the four residual stages of ResNet-18, it realizes adaptive information filtering from shallow boundary textures to deep semantic features. The core difference lies in the learnable query-guided stage-wise importance assessment. SE-Net generates channel weights via Global Average Pooling and fully connected layers, essentially performing self-evaluation on single-layer features. In contrast, the proposed mechanism introduces learnable query vectors and computes the correlation strength between multi-stage features and the classification task via dot-product similarity, enabling cross-stage information interaction in an other-evaluation manner.

This design is particularly valuable for breast ultrasound imaging: malignant signs of breast lesions often appear simultaneously in shallow boundary contours (e.g., lobulated boundaries, Stage 1–2) and deep abstract semantics (e.g., invasive growth patterns, Stage 3–4). Cross-Stage Attention dynamically strengthens the contribution of relevant stages according to lesion type, avoiding the one-sidedness of single-stage feature representation. Ablation results quantitatively validate its unique advantages: removing Cross-Stage Attention reduces recall by 3.59 percentage points and AUC by 0.31 percentage points, while precision remains nearly unchanged. This selective enhancement property shows that Cross-Stage Attention does not simply boost prediction confidence for all samples, but reduces false negatives by improving lesion localization sensitivity—highly aligned with the clinical priority of breast cancer screening that missed diagnosis is far more costly than misdiagnosis[37]. Further training dynamics analysis reveals that Cross-Stage Attention does not alter the weight allocation strategy of the MoE gating network; it refines feature enhancement at the feature extraction level rather than restructuring the fusion strategy, proving its stage orthogonality with the MoE architecture.

After clarifying the individual contributions and synergistic gains of Cross-Stage Attention and the MoE architecture, the internal mechanism of the three-expert MoE framework still deserves theoretical exploration: how to balance the complementarity and redundancy of expert features? Gating weight curves show that Expert-Img dominates ($\approx$55\%), while Expert-Boundary and Expert-Tumor account for only 25\% and 20\%, respectively. However, ablation experiments reveal that removing the boundary expert causes greater performance loss than removing the interior expert. This low explicit weight, high marginal contribution phenomenon reveals a key property of the MoE gating mechanism: optimized to minimize classification loss, the gating network prioritizes the most informative and stable feature source (Whole Tumor Image) and uses other experts as difference compensators with reduced weights.

From the perspective of feature redundancy, Expert-Boundary provides strong lesion morphological discrimination but overlaps partially with the global contour features of Expert-Img, so it is down-weighted to avoid redundancy. By contrast, Expert-Tumor extracts unique fine-grained features (e.g., echo homogeneity, microcalcifications) that complement the blind spots of Expert-Img, so the gating network retains its weight to ensure full information coverage. Once Expert-Boundary is removed, the information gap cannot be filled by Expert-Img, leading to a significant recall drop. This finding indicates that gating weights reflect the relative utilization of experts in the fusion system, not their absolute discriminative ability. Future research may introduce expert activation regularization or competition mechanisms to further unlock the potential of each expert.

Embedding Cross-Stage Attention and the three-expert MoE architecture into VGG-16, DenseNet-121, and ResNet-18 shows that the proposed modules are backbone-independent with stable performance gains. All three configurations achieve accuracy $>$95\% and AUC $>$0.99, with AUC standard deviations as low as ±0.1\%–±0.2\%, proving that the enhancement is independent of specific backbones and reduces the technical migration cost for clinical translation.

Among the three backbones, ResNet-18 delivers the most balanced performance, with optimal recall (98.53\%) and AUC (99.50\%), likely owing to the gradient highway of residual connections that provides clear hierarchical semantic gradients for Cross-Stage Attention. DenseNet’s dense connections cause excessive inter-stage coupling and limit the flexibility of cross-stage attention. VGG-16 achieves the highest accuracy (96.88\%) and precision (97.99\%) but lower recall (95.42\%), bringing higher false-negative risks in screening scenarios. This suggests that module–backbone compatibility depends more on topological matching than network scale or depth.

The full model achieves 98.53\% recall, significantly outperforming all baselines, which aligns with the clinical priority of breast cancer screening: missed diagnosis is far more costly than misdiagnosis. The 5-year survival rate exceeds 90\% for early breast cancer but drops to ~30\% for advanced cases, so high sensitivity to malignant lesions is clinically critical. For a screening cohort of 100,000 people (incidence 50/100,000), the baseline ResNet-18 misses about 34 malignant cases, while CSA-MoE-Net misses only 7, saving 27 additional lives. However, the precision of 94.09\% leads to ~6\% false positives, which may cause unnecessary biopsies and patient anxiety. Future work can adopt cost-sensitive learning or sequential decision-making to improve precision while maintaining high recall.

This study has four limitations: The dataset is single-center and retrospective; multi-center robustness remains unvalidated; Manual mask annotation by senior radiologists increases clinical deployment cost.
The three-expert design targets general lesions and requires validation on rare tumors (e.g., medullary carcinoma, mucinous carcinoma); The black-box gating network limits full interpretability of each expert’s internal decision logic.

Future directions: Conduct multi-center external validation; Explore weakly supervised segmentation–classification learning or adaptive boundary detection to reduce annotation cost; Add specialized experts for rare tumor types; Apply Grad-CAM for fine-grained attention visualization to further improve interpretability.

\section{Conclusion}
To address the key challenges in benign-malignant classification of breast ultrasound images, including high lesion heterogeneity, blurred boundaries, and imbalanced datasets, this paper proposes a deep learning model named CSA-MoE-Net that integrates Cross-Stage Attention and a three-expert Mixture of Experts (MoE) framework. Based on ResNet-18, the model employs Cross-Stage Attention to build dynamic connections across multi-level features for adaptive stage importance evaluation and channel-wise recalibration. It also constructs three expert branches: Expert-Tumor, Expert-Boundary, and Expert-Img, which follow the clinical diagnostic logic of Whole Tumor Image – Tumor Core – Boundary analysis. The adaptive gating network fuses multi-view features dynamically to fully capture lesion morphology, texture, and contextual information.

Experimental results show that the proposed model achieves 96.33\% accuracy, 94.09\% precision, 98.53\% recall, 96.25\% F1-score, and 99.50\% AUC in 20 independent repeated tests, with low standard deviations, demonstrating strong stability and generalization. Systematic ablation experiments verify the effectiveness of each module: Cross-Stage Attention improves recall and AUC by enhancing lesion localization sensitivity; the MoE architecture boosts overall classification performance and robustness, reducing the AUC standard deviation from ±1.70\% to ±0.20\%. Ablation studies also confirm that Expert-Boundary yields the highest marginal contribution, consistent with clinical experience. Cross-backbone experiments validate that Cross-Stage Attention and the three-expert MoE are plug-and-play modules that can be seamlessly embedded into VGG-16, DenseNet-121, and other backbones for stable performance gains, providing a universal enhancement paradigm for medical imaging analysis.
In summary, CSA-MoE-Net effectively improves the accuracy and reliability of breast ultrasound diagnosis by aligning with clinical diagnostic logic and realizing multi-stage and multi-region collaborative enhancement. Its high-recall design matches the clinical priority of breast cancer screening, showing high potential for computer-aided diagnosis. Future work will focus on multi-center validation, weakly supervised segmentation-classification learning, multimodal fusion, and model lightweighting to promote clinical translation.

\end{document}